\documentclass{article}
\usepackage[utf8]{inputenc}
\usepackage{bm}
\usepackage{hyperref}       
\usepackage{url}            
\usepackage{booktabs}       
\usepackage{amsfonts}       
\usepackage{nicefrac}       
\usepackage{microtype}      
\usepackage{xcolor}         
\usepackage{graphicx}
\usepackage{caption}
\usepackage{float}
\usepackage{subcaption}
\usepackage{gensymb}
\usepackage{xcolor}
\usepackage{amsmath}
\usepackage{placeins}
\usepackage{selectp}
\usepackage[ruled,linesnumbered]{algorithm2e}
\usepackage[numbers]{natbib}
\newcommand{\diag}{\mathop{\mathrm{diag}}}

\usepackage{authblk}

\title{\textbf{\emph{Efficient Multi-Prize Lottery Tickets}: Enhanced Accuracy, Training, and Inference Speed}}

\author[1]{Hao Cheng}
\author[1]{Pu Zhao}
\author[1]{Yize Li}
\author[1]{Xue Lin}
\author[2]{James Diffenderfer}
\author[2]{Ryan Goldhahn}
\author[2]{ Bhavya Kailkhura}
\affil[1]{Northeastern University}
\affil[2]{Lawrence Livermore National Laboratory}
\date{}

\begin{document}

\maketitle
\begin{abstract}
Recently, Diffenderfer \& Kailkhura \cite{diffenderfer2021multiprize} proposed a new paradigm for learning compact yet highly accurate binary neural networks simply by pruning and quantizing randomly weighted full precision neural networks. However, the accuracy of these multi-prize tickets (MPTs) is highly sensitive to the optimal prune ratio, which limits their applicability. Furthermore, the original implementation did not attain any training or inference speed benefits. In this report, we discuss several improvements to overcome these limitations. We show the benefit of the proposed techniques by performing experiments on CIFAR-10.
\end{abstract}
\section{Introduction}
It is found that randomly-initialized dense networks contain subnetworks that  can  achieve test accuracy comparable to the trained dense network \cite{ramanujan2020s}. Based on this, the authors in \cite{diffenderfer2021multiprize} propose (and prove) a stronger Multi-Prize Lottery Ticket Hypothesis:

\noindent \textit{ A sufficiently over-parameterized neural network with random weights contains
several subnetworks (winning tickets) that (a) have comparable accuracy to a
dense target network with learned weights (prize 1), (b) do not require any further training to achieve prize 1 (prize 2), and (c) is robust to extreme forms of
quantization (i.e., binary weights and/or activation) (prize 3).}

This provides a new paradigm for learning compact yet highly accurate binary
neural networks simply by pruning and quantizing randomly weighted full precision neural networks. The proposed   algorithm can find multi-prize tickets (MPTs)  with SOTA accuracy for binary neural networks without ever updating the weight values. 

The high accuracy of subnetworks can be achieved at a specific sparsity ratio. Since the model weights are not trained, if the sparsity ratio is not appropriately  configured, the accuracy performance may suffer.  
Besides, although the model weights are not updated, the MPTs algorithm needs to train the scores (each score corresponding to one weight)   with moderate training efforts. 
Moreover, during inference, the sparsity of the subnetwork is not fully exploited for efficient inference so the inference speed is still the same as the full dense model without any acceleration effects.

To address the above issues, we further consider how to improve the MPTs algorithm on these fronts. We propose four improvements to enhance the accuracy, training, and inference speed. Our improvements include:
\begin{itemize}
	\item  Power-propagation \cite{schwarz2021powerpropagation} parameterization of scores to improve the accuracy performance on a wider range of sparsity ratios, 
 \item Fine-tuning the model to further improve the accuracy,
	\item  Thresholding instead of sorting-based pruning to accelerate the training, and 
	\item  Structured pruning to remove kernels and improve the inference speed. 
\end{itemize}

Next, we discuss these improvements in detail and provide empirical support to justify their superiority over the original MPT method\footnote{Although \cite{diffenderfer2021multiprize} supports both binary weight and binary activation MPTs, we only focus on binary weight MPTs in this report for simplicity.}.

\section{Incorporating Power-Propagation into  MPTs}

\subsection{Incorporating Power-Propagation Parameterization}

We incorporate Power-propagation \cite{schwarz2021powerpropagation} into MPTs \cite{diffenderfer2021multiprize}. Different from other model training methods, MPT does not train the model weights. Instead, it keeps the model weights untouched as their initial random values.  Note that we only focus on the binary weight instead of binary activations. For each weight, it needs to determine whether to prune or binarize this parameter.  
MPT uses a score for  each model  weight to indicate its importance. Thus, it trains and sorts the scores to determine which weights should be pruned and which weights should be preserved.   
We can see that score updating plays an important role in MPTs.
Thus, we adopt Power-propagation parameterization to further improve the score updating performance.    Specifically, the original score $\bm S$ is replaced with  a mapping function $\Psi(\bm s) = \bm s|\bm s|^{\alpha-1} = \bm S $.  We only apply this transformation to the scores in MPT, leaving other parameters untouched.    With the  loss  $\mathcal{L}(\Psi(\bm s))$, the gradients w.r.t. to $\bm s$ can be obtained as follows,
\begin{equation}
    \frac{\partial \mathcal{L}( \Psi(\bm s))}{\partial \bm s}= \frac{\partial \mathcal{L}}{\partial \Psi(\bm s)}  \frac{\partial \Psi(\bm s)}{\partial \bm s} =  
    {\frac{\partial \mathcal{L}}{\partial \Psi(\bm s)}}{\diag(\alpha |\bm s|^{\circ\alpha-1})},
    \label{eq:powerprop_update}
\end{equation}
where $\diag$ denotes a diagonal matrix, and $|\bm s|^{\circ \alpha-1}$ indicates computing the   power $\alpha -1$ of $|\bm s|$ element-wise. 
 $ \frac{\partial \mathcal{L}}{\partial \Psi(\bm s)} $ is the derivative w.r.t. to the original score $\bm S = \bm s|\bm s|^{\circ\alpha-1}$. 
In our update, it is additionally multiplied (element-wise) by a  factor $ {\alpha |\bm s|^{\alpha-1}}$, which   scales the step taken proportionally to the magnitude of  each entry. Note that this update is different from simply scaling the gradients  of $\bm S$ by the magnitude of the parameter (raised at $\alpha-1$), since the update is applied to $\bm s$ not $\bm S$, and is scaled by $\bm s$.  The update rule \eqref{eq:powerprop_update} has the following properties:
\begin{itemize}
    \item[(i)] If $\alpha>1$,  $\frac{\partial \mathcal{L}}{\partial \bm s} = 0$ whenever $\bm s=0$ due to the $\alpha |\bm s|^{\alpha-1}$ factor.
    \item[(ii)] In addition, 0 is surrounded by a plateau and hence scores are less likely to change sign (gradients become vanishingly small in the neighborhood of 0 due to the scaling). This should negatively affect initialization which allows for both negative and positive values, but it might have bigger implications for biases.
    \item[(iii)] This update is naturally obtained by the backpropagation algorithm since backpropagation implies applying the chain-rule from the output towards the variable of interest, and Eq. \eqref{eq:powerprop_update} simply adds another composition (step) in the chain before the variable of interest. 
\end{itemize}

\subsection{Performance of Power-Propagation MPTs}

\begin{figure}[th]
\centering
\includegraphics[width=0.45\textwidth]{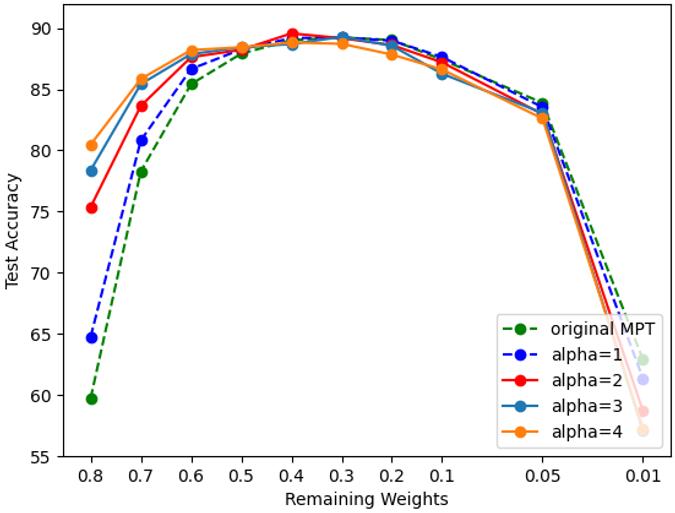}
\includegraphics[width=0.45\textwidth]{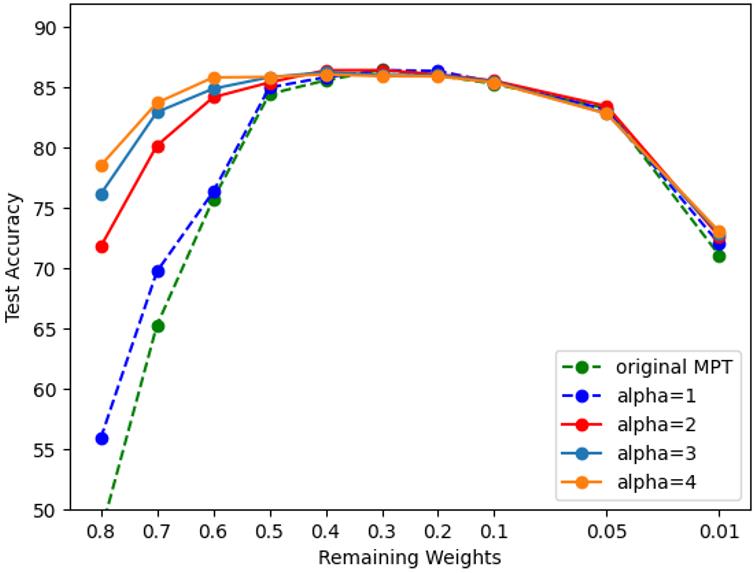}
\caption{Performance of Power-Propagation. The left figure shows the results of CONV-6 with 6 CONV layers and the right figure shows the results of CONV-4 with 4 CONV layers. }
\label{fig: power_performance}
\end{figure}

We  show the performance of MPTs with power-propagation in Fig. \ref{fig: power_performance}. We conduct experiments on CIFAR-10 for CNNs  with varying depths (including CONV-4 with 4 CONV layers, CONV-6 with 6 CONV layers and CONV-8 with 8 CONV layers). We show the results of different hyperparameter $\alpha$ values ($\alpha = 1, 2, 3, \text{or}, 4$). Note that when $\alpha=1$, it is just the case of the original score $\bm S$ training without power-propagation. We adopt the default hyperparameter setting from the original MPT paper to train the model \footnote{Note that the code support both global and layerwise pruning. Since global pruning usually performs better,  we adopt global pruning as our main focus.}.

In Fig. \ref{fig: power_performance}, we can observe that power-propagation can increase the accuracy performance when the pruning ratio is small (smaller than 0.5) with less pruned weights.  When the pruning ratio is above 0.5, the accuracy performance is similar to the original case without power-propagation. When $\alpha=1$, the performance is very similar to the original case without power-propagation, demonstrating the equivalence of $\alpha=1$ and no power-propagation. When $\alpha$ increases,  the accuracy performance under small pruning ratios also improves.

As shown in Fig. \ref{fig: power_performance},  since power-propagation can help to improve the accuracy when the pruning ratio is small,  the performance of MPT becomes more robust with  high accuracy performance achieved on a wider range of sparsity ratios.

\subsection{Score distribution with Power-Propagation}

\begin{figure}[th]
\centering
\includegraphics[width=0.45\textwidth]{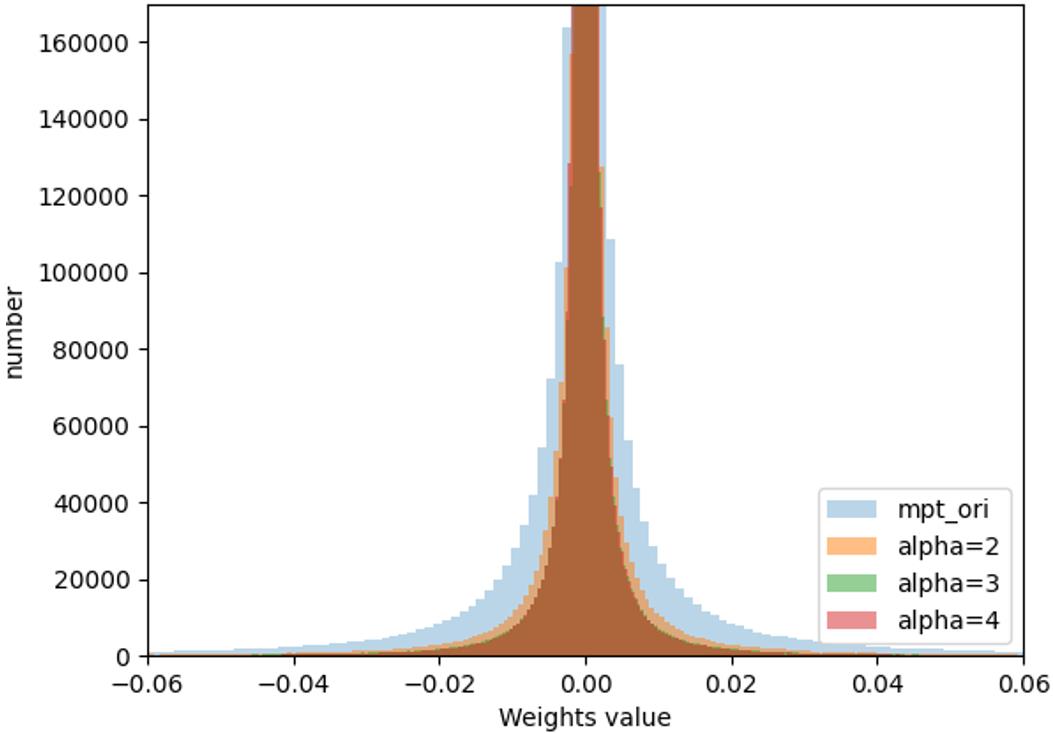}
\includegraphics[width=0.44\textwidth]{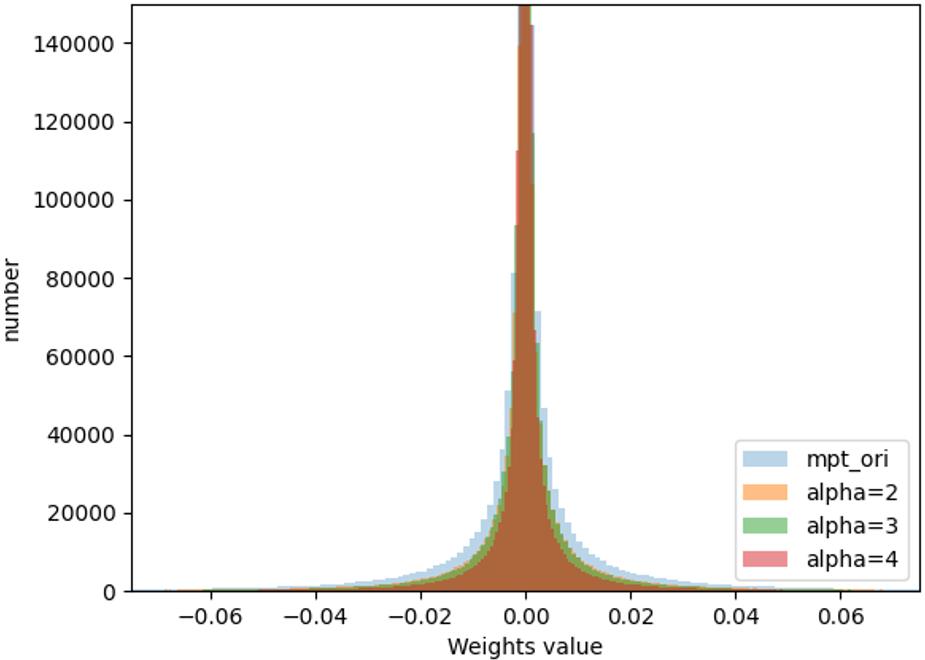}
\caption{Distribution of scores for Power-Propagation. The left figure shows the distribution CONV-8 with 8 CONV layers. The right figure shows the distribution of CONV-6 with 6 CONV layers.  }
\label{fig: power_distribution}
\end{figure}

To explore why power-propagation can improve the MPT performance, we  show the distribution of scores of certain convolution layer in the model in Fig. \ref{fig: power_distribution}. We can see that after adopting power-propagation, the integral value range of the score  shrinks. It means that the  scores are distributed in a  smaller value range. Therefore, updating the scores becomes more efficient and a small change can effectively update the status of pruned or preserved.

\section{Incorporating Weights Finetuning in MPTs}

\subsection{ Finetuning Method}

In the original MPTs, the weights are not trained and are kept as randomly initialized values. MPT assigns a score for each weight and trains the scores to determine whether to prune or binarize the weights. In this section, we further consider whether finetuning the weights can further improve the performance \cite{DBLP:journals/corr/abs-2110-09510,Li2020Rethinking}.  Specifically, we first use MPTs to obtain a sparse binarized model.   This model can be represented by  $\bm W_b \cdot \bm M$, where $\bm W_b$ is the binarized weights and $\bm M$ is the mask with 0 or 1 values to determine whether to prune each weight. Based on MPTs, we have the following,
\begin{align}\label{eq:mask}
\alpha^{(j)} & = \| \bm{M}^{(j)} \odot \bm{W}^{(j)} \|_{1} / \| \bm{M}^{(j)} \|_{1}  \\ 
\bm W_b^{(j)} & = \alpha^{(j)} \cdot   \text{sign}(\bm{W}^{(j)})
\end{align}
where $\bm W^{(j)}$ is the original weights  in the $j$-th layer. After we run MPTs, we can obtain  $\bm W_b$ and $\bm M$.  
Next, when we start to finetune the MPTs model, we keep the mask $\bm  M$ fixed and only update the model weights $\bm W$ with gradients.  Note that different from MPTs where the gradients are computed for scores to update mask, here the gradients are computed for model weights. Besides, although model weights are updated, during the inference, they are still transformed into binary values following Eq. \eqref{eq:mask}.

\subsection{Performance of Finetuning}

\begin{figure}[ht]
\centering
\includegraphics[width=0.32\textwidth]{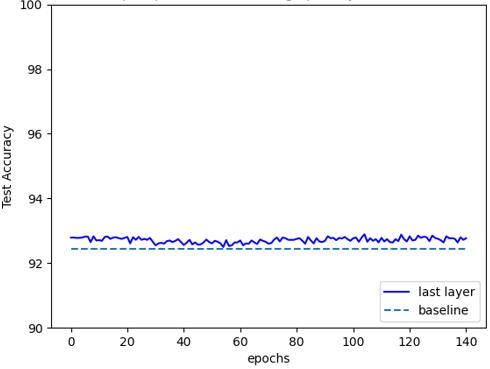}
\includegraphics[width=0.32\textwidth]{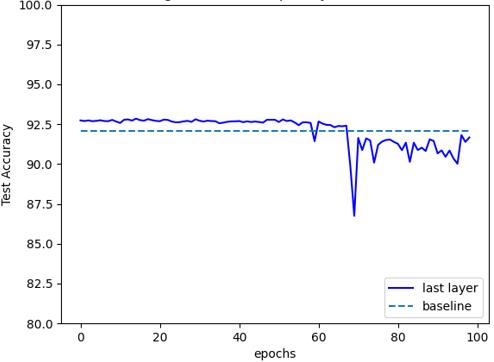}
\includegraphics[width=0.32\textwidth]{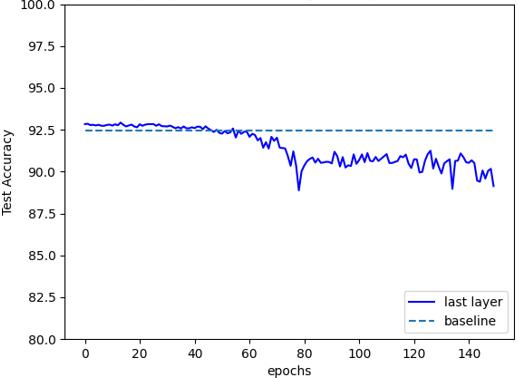}
\caption{Performance of Finetuning. The three figures from left to right show the results of finetuning the first layer, the last layer, and the whole model.}
\label{fig: finetuning_three}
\end{figure}

We  show the performance with weights finetuning in  Fig. \ref{fig: finetuning_three}. We conduct experiments on CIFAR-10 with Resnet-18.  For the finetuning, we try three cases including finetuning the first layer, the last layer, or the full model with all layers. 

In Fig. \ref{fig: finetuning_three}, we report the performance of the three cases discussed above.  
We can see that  finetuning can help to slightly improve the accuracy performance in all three cases. Usually finetuning the last layer can obtain the largest accuracy  improvement. However, we should not finetune for too many epochs to avoid the accuracy drop. Usually, the finetuning epochs can be smaller than 50.  


For the funetuning, we perform a grid search over different hyperparameters summarized in Table \ref{tab:finetune}. The best accuracy performance is achieved with the SGD optimizer, 0.001 learning rate with cosine scheduler, 256 batch size, and finetuning the last layer, which is highlighted in Table \ref{tab:finetune}. 

\begin{table}[t]
\centering
\begin{tabular}{l|l|l|l|l}
\hline
   Optimizer       & \textcolor{red}{SGD}  & Adam  &   &  \\ \hline
lr scheduler   & multi-step & \textcolor{red}{cosine} & constant & \\ \hline
lr   &  0.1 & 0.01 & \textcolor{red}{0.001} & 0.0001   \\ \hline
Batch size & 64 & 128 & \textcolor{red}{256} & 512 \\ \hline
Fine-tune &  full model & \textcolor{red}{last layer} & first layer &  \\ \hline
\end{tabular}
\caption{Hyper-parameter settings for funetuning.}
\label{tab:finetune}
\end{table}

\section{Replacing sorting with thresholding  to accelerate the training}

\subsection{ Thresholding Method}

Original MPT implementation requires sorting of the scores to determine top weights that are important and should be preserved.  The sorting operation is usually time-consuming, especially for the DNN layers with  a large number of parameters.  Thus, to accelerate the training, we adopt a thresholding method to replace the sorting operation. Specifically, once the scores are updated, we do not sort the scores. Instead, we directly compare the scores with a predefined threshold. If the score is larger than the threshold, the corresponding weight is preserved. Otherwise, the corresponding weight is pruned.  
Thus, we do not need to sort the scores in each iteration during training and the training speed can be accelerated effectively.

\subsection{Performance of Thresholding Based MPT}

\begin{figure}[h]
\centering
\includegraphics[width=0.65\textwidth]{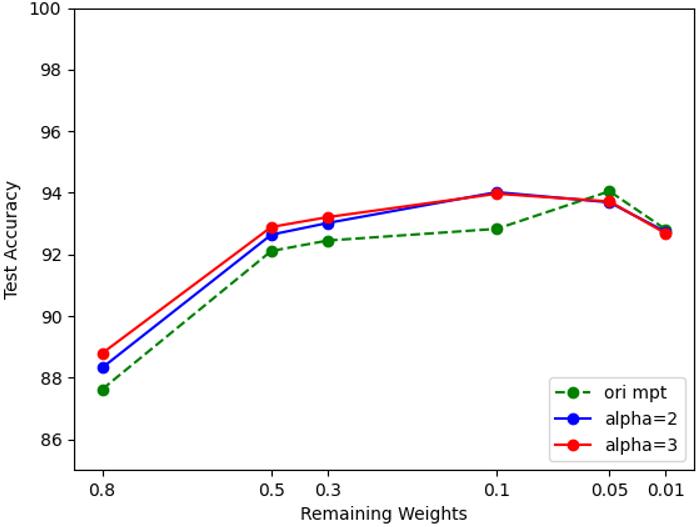}
\caption{Performance of thresholding-based MPT.}
\label{fig: threshold}
\end{figure}

We conduct experiments on CIFAR-10 with Resnet-18 and show the results of thresholding in Table~\ref{tab1} and \ref{tab2}.  We adopt the default hyperparameter setting in the original MPTs paper to train the model.  We can observe that the training time for each epoch can be reduced by 13\%. Here we  incorporate the power-propagation into the MPTs for efficient score training.  We also show the accuracy performance with thresholding in Fig. \ref{fig: threshold}. We can see that the accuracy can be even higher than the original MPT implementation. 

\begin{table}[t]
\centering
\begin{tabular}{l|l|l|l}
\hline
          & $\alpha=1$  & $\alpha=2$  & $\alpha=3$  \\ \hline
Sorting   & 24.3 s/epoch & 25.4 s/epoch & 26.8 s/epoch \\ \hline
Thresholding & 21.5 s/epoch & 22.3 s/epoch & 24.5 s/epoch \\ \hline
\end{tabular}
\caption{Training time comparison between sorting and thresholding.}
\label{tab1}
\end{table}

\begin{table}[t]
\centering
\begin{tabular}{l|l|l|l}
\hline
          & $\alpha=1$  & $\alpha=2$  & $\alpha=3$  \\ \hline
Sorting   & 4.05 min & 4.23 min & 4.47 min \\ \hline
Thresholding &  3.58 min & 3.73 min & 4.07 min \\ \hline
\end{tabular}
\caption{Training time comparison for 10 epochs between sorting and thresholding.}
\label{tab2}
\end{table}

\section{Structured pruning to accelerate the inference}

\subsection{ Posthoc Structured Pruning Method}

\begin{figure}[h]
\centering
\includegraphics[width=0.75\textwidth]{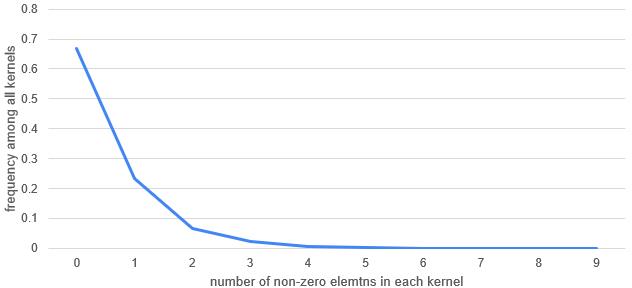}
\caption{Distribution of sparse kernels of all layers.}
\label{fig: sparsity}
\end{figure}

Although MPTs use unstructured pruning for training, we find that, in the case with a high pruning ratio, there is a certain sparsity structure that can be utilized to accelerate the inference. For example, if most weights are pruned, there are also many all-zero $3\times 3$ convolutional (CONV) kernels in each CONV layer.  To demonstrate this, we plot the number of zero kernels in each layer for a MPT model of 95\% pruning ratio in Fig. \ref{fig: sparsity}.  We can see that there are 67\% kernels that are all zeros among all layers. This implies that we can exploit the kernel sparsity to accelerate the training \cite{zhong2022revisit,9321463,DBLP:journals/corr/abs-1909-05073}.

Specifically, if certain kernels are all zeros, meaning the whole kernel can be pruned, we can try to remove the whole kernel in the computation graph, thus avoiding the computations related to this pruned kernel. The acceleration rate can be computed as 
\begin{align}
AR = \frac{\sum N \times M \times D \times D \times k \times k}
{\sum P \times D \times D \times k \times k}
\end{align}
where $N$ and $M$  are the input and out channel numbers in one layer, $D$ is the size of the feature map and $k$ is the kernel size of that layer. The sum is taken over all layers. The acceleration rate is the rate between the original  computation counts and the  sparse computation counts after removing all-zero kernels, which shows the change of the computation number, and the corresponding inference speed. 

\subsection{Performance of Structure Pruning}

We  show the performance with structure pruning in Table \ref{tab3}. We conduct experiments on CIFAR-10 with Resnet-18. The default hyperparameter setting  in the original MPTs paper \cite{diffenderfer2021multiprize} is adopted to train the model.  We can see that by exploring the structure sparsity in the MPT model, we can remove the all-zero kernels and accelerate the inference effectively. 

\begin{table}[t]
\centering
\begin{tabular}{l|l|l|l}
\hline
     Sparsity      & 70\%  & 90\% & 95\% \\ \hline
Acceleration rate   &  1.4$\times$  &  2.3$\times$  & 3.0$\times$  \\ \hline
\end{tabular}
\caption{Acceleration rate of various sparsities.}
\label{tab3}
\end{table}

\section{Conclusion}

We  proposed various enhancements for making MPTs more efficient. Our results demonstrated that power-propagation and finetuning can help to improve the accuracy performance. Further, the thresholding can accelerate the MPT training and the structure pruning can accelerate the inference speed. 

\section*{Acknowledgement}
This work was performed under the auspices of the U.S. Department of Energy by the Lawrence Livermore National Laboratory under Contract No. DE-AC52-07NA27344 and was supported by LLNL-LDRD Program under Project No. 20-SI-005.

\clearpage
%
%
\bibliographystyle{splncs04}
\bibliography{egbib}

\end{document}